\documentclass[runningheads,capitalise,capitalize]{llncs}
\usepackage[T1]{fontenc}
\usepackage{graphicx}
\usepackage{amsmath,amssymb}
\usepackage{booktabs}
\usepackage{multirow}
\usepackage{xcolor}
\usepackage{enumitem}
\usepackage{comment}
\usepackage{wrapfig}
\usepackage{eccv}
\usepackage[pagebackref,breaklinks,colorlinks,citecolor=eccvblue]{hyperref}
\usepackage{eccvabbrv}
\usepackage{orcidlink}
\usepackage{amsmath,amssymb}
\usepackage{graphicx}
\usepackage{booktabs}
\makeatletter
\let\llncssubparagraph\subparagraph
\let\subparagraph\paragraph
\makeatother

\usepackage{titlesec}

\makeatletter
\let\subparagraph\llncssubparagraph
\makeatother
\usepackage[accsupp]{axessibility}  % Improves PDF readability for those with disabilities.

\usepackage{xcolor}
\usepackage{soul}
\sethlcolor{yellow}

\usepackage{caption}
\newcommand{\cls}{\texttt{[CLS]}}
\newcommand{\R}{\mathbb{R}}
\newcommand{\Ahat}{\hat{A}}

\newcommand{\sa}[1]{\textcolor{black}{#1}}
\title{Fragment-Aware Vision Transformers for Fresco-Fragment Style Classification}
\titlerunning{Fresco-Fragment Style Classification}

\author{Sara Miketek\inst{1}\orcidlink{0009-0009-5959-3318} \and
Biagio Barchielli\inst{1}\orcidlink{0009-0001-1711-9759} \and
Nadeem Iqbal Kajla\inst{2}\orcidlink{0000-0001-8722-7790}
\and
\\ Sinem Aslan\inst{3}\orcidlink{0000-0003-0068-6551}}

\authorrunning{S. Miketek et al.}
\institute{Ca' Foscari University of Venice, Venice, Italy \and
Dundalk Institute of Technology, Dundalk, Ireland
\and
University of Milan, Milan, Italy\\
\email{sinem.aslan@unimi.it}}
\date{}
\begin{document}
\maketitle
%

% =====================================================================
\begin{abstract}
% CHANGE: abstract reframed. Kept Nadeem's framing and the CLEOPATRA story, but
% (a) the headline is now the simple ensemble, (b) the graph is reported as a
% secondary comparison, and (c) interpretability is added as a core component.
Artistic style classification is usually studied on complete artworks, where
models can exploit global composition, spatial organisation, and iconographic
structure. In archaeological settings, however, artworks often survive only as
fragmented remains, forcing recognition from incomplete, irregular, and
context-limited visual evidence. We study fresco-fragment style classification
using a progressive transformer-based framework. Starting from a ViT-B/16
baseline, we introduce foreground-guided masking to suppress background-only
tokens, inpainting-based geometric regularisation to align irregular fragment
supports with the ViT patch grid, and a supervised contrastive objective
that operates on predictive distributions through a \sa{Kullback--Leibler}
similarity and consistently improves every branch. We combine the branches with
a deliberately simple learnable logit ensemble. Experiments on CLEOPATRA and
POMPAAF show that fragment-aware modelling improves over the standard ViT
baseline, with the ensemble increasing accuracy from $0.604$ to $0.656$ and macro-F1 from $0.596$ to $0.648$ on CLEOPATRA, and outperforming %outperforms 
the best single branch in
four of six fragmentation settings on POMPAAF. 
%We additionally evaluate a more complex graph-fusion variant and find it does not consistently outperform the simple ensemble. 
We additionally evaluate a more complex graph-fusion variant and find that it
matches the simple ensemble on POMPAAF while offering only a small,
dataset-specific gain on CLEOPATRA, which does not justify its added complexity. Beyond these empirical gains, our contribution is twofold:
a distribution-level contrastive objective that consistently sharpens
single-branch recognition, and an interpretability analysis that verifies the
models exploit genuine painted evidence, while quantifying that the inpainting-based branch draws part of its attribution from the synthesised surround. %while exposing a synthesis-driven shortcut in the inpainting-based branch.
\end{abstract}

% =====================================================================
\titlespacing*{\section}{0pt}{1.0ex plus 0.2ex minus 0.2ex}{0.6ex}
\titlespacing*{\subsection}{0pt}{0.8ex plus 0.2ex minus 0.2ex}{0.4ex}
\titlespacing*{\subsubsection}{0pt}{0.6ex plus 0.2ex minus 0.2ex}{0.3ex}

\section{Introduction}

\label{sec:intro}
% CHANGE: Nadeem's introduction kept almost verbatim; only the interpretability
% motivation and the contributions list are new.

Artistic style classification is usually studied on complete artworks, where
models can exploit global cues such as composition, spatial organisation,
subject matter, and colour distribution \cite{van2015toward}. This assumption
often fails in archaeological and cultural-heritage settings, where artworks
survive as fragments and style must be inferred from incomplete, irregular, and
context-limited evidence. In this setting, local cues such as texture, pigment
distribution, brushwork, decorative motifs, and surface patterns become central.
Such fragment-level recognition is relevant to archaeological reconstruction,
where stylistic information can help organise fragments before reassembly
\cite{cascone2023classification}.

Recent fragment-level benchmarks such as CLEOPATRA
\cite{cascone2023classification} and POMPAAF \cite{elkin2025recognizing} show
that style recognition from fragments is feasible, but also expose challenges
that are less prominent in complete-image classification: irregular supports,
background-only regions introduced by resizing, and the need to aggregate local
stylistic cues without full composition.

Vision Transformers are a natural framework for this problem because they
represent images as patch tokens and model their interactions through
self-attention \cite{dosovitskiy2020image}. However, fragmented inputs raise two
specific issues: many tokens may correspond only to background, and irregular
fragment boundaries do not align well with the regular ViT patch grid. We
therefore study foreground-guided token masking, motivated by prior work on
reducing uninformative tokens \cite{liang2022not,cheng2024putting}, and
inpainting-based geometric regularisation, used here only to regularise geometry
rather than to reconstruct the artwork
\cite{criminisi2004region,suvorov2022resolution}. We also test whether explicit
patch-relation modelling adds value beyond simple branch fusion
\cite{chen2024unified,kim2024rethinking}. Because heritage applications require
predictions to be tied to meaningful visual evidence, we treat interpretability
as a core part of the evaluation.
 
% CHANGE: roadmap sentence restored (condensed from Nadeem's lines 092-102), to
% keep the narrative thread baseline -> masking -> inpainting -> ensemble.

In this work we study fresco-fragment style classification through a
progressive framework that moves from a ViT-B/16 baseline to foreground masking,
inpainting-based geometric regularisation, and a simple learnable ensemble, and
we analyse every variant with post-hoc explanations to identify which fragment
evidence drives its decisions. \sa{Our central message is that, in this fragmented setting, the leverage lies less in architectural complexity than in understanding what the models actually use, i.e., a simple ensemble matches a far more complex graph variant, while an interpretability analysis confirms that the models rely on genuine painted evidence and quantifies how the inpainting-based branch additionally draws part of its evidence from the synthesised surround. } % and reveals that inpainting introduces a synthesis-driven shortcut. 

Our contributions are listed as\footnote{Code is available at \url{https://github.com/skippa1da2flippa/fragment_classification}.}:
\begin{itemize}[leftmargin=1.3em,itemsep=2pt] 
  \item We investigate ViT-based style classification under visual fragmentation,
        focusing on archaeological fresco fragments where stylistic evidence is
        incomplete and spatially irregular.
  \item We evaluate two complementary fragment-aware strategies, namely, foreground-guided token masking and geometry regularisation through inpainting, against a
        standard ViT baseline, and a %corrected 
        \sa{KL contrastive}
        objective.
\item We combine the branches with a simple learnable logit ensemble, and
        report that a more complex graph-fusion variant offers only a small,
        inconsistent gain over it; we therefore adopt the simpler fusion on
        grounds of parsimony and treat the graph variant as secondary.        
  \item We provide \sa{an interpretability analysis}  %interpretability study
  (region-level foreground-relevance analysis, attention rollout, SHAP-BPT, controlled ablations) that
        characterises what each branch attends to and validates the masking design, and quantifies that the inpainting-based branch draws part of its evidence from synthesised content. %,        and exposes a synthesis-driven shortcut in the inpainted branch.
\end{itemize}

% =====================================================================
% =====================================================================
\section{Related Work}
\label{sec:related}
% CHANGE: reverted to Nadeem's fuller related work. Only two targeted edits:
%   (a) the graph-ViT paragraph is trimmed and reframed toward the demotion;
%   (b) the explainability paragraph is strengthened (not shortened), since
%       interpretability is now a core contribution. All else is Nadeem's prose.

\paragraph{Artistic style classification.}
Computational style analysis has often been studied under the assumption that the
complete artwork is available. In this setting, models can exploit global
composition, object layout, colour distribution, and spatially localised visual
evidence. Van Noord et al.  \cite{van2015toward} showed that convolutional models
can learn discriminative artist-specific patterns from digitised artworks and used
sensitivity maps to identify the regions that support a prediction. More recent
work has also examined how visual models separate stylistic and semantic cues,
including studies of cross-attention behaviour in text-to-image
models \cite{ferrara2025cow}. These works show that style can be represented
computationally, but they mainly operate on complete or globally observable
images. This differs from archaeological material, where the visual field is often
incomplete and the model cannot rely on full composition.

\paragraph{Style recognition from fragments and archaeological material.}
Fragment-level style recognition is closer to the conditions faced in
cultural-heritage reconstruction. Cascone et al. \cite{cascone2023classification}
introduced the CLEOPATRA dataset and formulated artistic-style recognition as a
classification problem over fragments, showing that style information can remain
available even when the artwork is divided into pieces. CLEOPATRA, however, remains
a simulated setting: fragments are generated from complete source images, so the
archaeological difficulties of erosion, missing material, and severe geometric
irregularity are only partially represented. Elkin et al. \cite{elkin2025recognizing}
moved the problem further into the archaeological domain, studying style
recognition from archaeological image fragments and introducing the POMPAAF
benchmark of Pompeian wall-painting fragments generated under different
fragmentation schemes. The RePAIR benchmark \cite{tsesmelis2024re} provides a
complementary, more realistic perspective, focusing on real Pompeian fresco
fragments with irregular shapes, erosion, missing material, and multimodal
2D--3D information; although it is designed for reassembly rather than style
classification, it documents the practical difficulty of fragment analysis in
realistic archaeological settings. Together, these studies motivate fragment-aware
recognition methods that can operate without complete composition or stable object
structure.

 \paragraph{Vision Transformers and fragment-aware token processing.}
Vision Transformers represent an image as a sequence of patch tokens and use
self-attention to model interactions among them \cite{dosovitskiy2020image}.
Subsequent variants improved this formulation through data-efficient training,
hierarchical windowed attention, deeper architectures, and multi-scale patch
processing \cite{touvron2021training,liu2021swin,touvron2021going,chen2021crossvit}.
These properties make transformer models suitable for fragment analysis, where
stylistic evidence may appear in local patches rather than in the full image.
However, fragmented inputs introduce a specific problem: many tokens may correspond
to background introduced by irregular fragment geometry rather than to painted
surface. Liang et al. \cite{liang2022not} showed that not all ViT patches
contribute equally and proposed token reorganisation to reduce uninformative
computation, while Cheng et al. \cite{cheng2024putting} used
foreground--background masked attention to separate object evidence from background
regions. These works support the use of foreground-guided masking in fragment
classification, where the valid painted region occupies only part of the square
input canvas.

\paragraph{Geometry regularisation through inpainting.}
A second way to handle irregular fragments is to regularise the input geometry
before classification. Exemplar-based inpainting fills missing regions by copying
information from known image areas according to patch
priorities \cite{criminisi2004region}. More recent large-mask inpainting methods
use Fourier convolutions to model long-range spatial dependencies and improve
completion under large missing regions \cite{suvorov2022resolution}. In the present
setting, inpainting is not treated as an archaeological reconstruction of the
original fresco; instead, it is used as a preprocessing strategy to reduce the
mismatch between irregular fragment boundaries and the regular patch grid assumed
by Vision Transformers.

\paragraph{Graph-based fusion and explainability.}
% CHANGE (a): this graph-ViT paragraph was longer and more enthusiastic in
% Nadeem's draft; trimmed and reframed toward the demotion (purple = changed).
Since global composition is largely unavailable in fragments, relations among
local patch representations may, in principle, provide additional evidence for
style recognition. Several works have connected Vision Transformers with
graph-based relational modelling: Chen et al.~\cite{chen2024unified} proposed a
relational graph view of ViTs, Devaguptapu et al.~\cite{devaguptapu2024semantic}
regularised self-supervised ViTs through semantic graph consistency, and Kim and
Ko~\cite{kim2024rethinking} reconsidered attention in ViTs using graph structures;
related hybrid models combine transformer features with graph attention or graph
convolution in other visual and multimodal
domains~\cite{venkatraman2025sag,fixelle2025hypergraph,zhang2026gc,dogga2026hybrid,jia2025multimodal,tran2025novel}.
%These results motivate testing graph-based fusion in our setting; however, as we report in \cref{sec:res-pomp}, a graph variant does not consistently outperform a simple learnable logit ensemble, so we treat relational fusion as a secondary component rather than a central one.
These results motivate testing graph-based fusion in our setting; however, as
we report in \cref{sec:res-pomp}, a graph variant offers no consistent advantage
over a simple learnable logit ensemble relative to its added complexity, so we
treat relational fusion as a secondary component rather than a central one.

% CHANGE (b): explainability paragraph STRENGTHENED (interpretability is now a
% core contribution), not shortened. Purple = added/strengthened.
Explainability is also central in cultural-heritage applications, where predictions
should be tied to meaningful visual evidence. General attribution methods include
LIME~\cite{ribeiro2016should}, SHAP~\cite{lundberg2017unified}, and
Grad-CAM~\cite{selvaraju2017grad}. For transformers, attention
rollout~\cite{abnar2020quantifying} and relevance-propagation
methods~\cite{chefer2021transformer} provide token-level explanations beyond raw
attention visualisation, while for graph models GNNExplainer~\cite{ying2019gnnexplainer}
and PGExplainer~\cite{luo2020parameterized} identify influential nodes, edges, or
subgraphs. SHAP-BPT~\cite{rashid2026shapbpt} introduces data-aware binary partition
trees for image explanations, aligning attributions with image morphology rather
than treating pixels as an unstructured set. These tools are particularly
relevant here because fragment classification requires not only high accuracy but
also evidence that the model relies on painted foreground rather than silhouette,
background, or synthesis artefacts. Unlike prior art-analysis work, we do not use
these methods only for qualitative visualisation; we combine attention rollout and
SHAP-BPT with a quantitative foreground-relevance measure and a controlled
input-ablation study to test, rather than assume, which visual evidence drives each
branch.

\paragraph{Limitations of existing work.}
% CHANGE: Nadeem's "limitation" paragraph kept; closing sentence updated (purple)
% to match the demoted-graph / interpretability framing.
Existing work establishes that artistic style can be learned from images and that
fragment-level recognition is feasible. However, the specific behaviour of
patch-based transformer models under irregular fragment geometry remains
insufficiently studied. Prior work does not systematically compare standard ViT
classification, foreground-guided token masking, and inpainted geometric
regularisation for fresco-fragment style recognition, nor does it verify,
quantitatively, whether such models rely on genuine painted evidence rather than
on fragment shape or synthesised content. This work addresses that gap by
evaluating these strategies within a common fragment-level classification
framework and analysing them with quantitative interpretability tools.
% =====================================================================
% =====================================================================
% ==============   SECTIONS 3 ONWARD (revised on Nadeem)   =============
% =====================================================================

\section{Proposed Methodology}
\label{sec:method}
% Nadeem's overview kept; only the last clause changed (graph -> simple ensemble).
Given a fresco fragment image, the goal is to predict its artistic style from
incomplete and spatially irregular visual evidence. The proposed framework is
organised in stages. First, a standard ViT-B/16 model is used as a baseline.
Second, a masked ViT variant suppresses patch tokens that correspond only to
background. Third, inpainting is used as a geometric regularisation strategy for
irregular fragments. Finally, the three trained transformer variants are
combined through a simple learnable logit ensemble, and we additionally test a
graph-based fusion variant that we report as a secondary comparison.
 
\subsection{Notation}
\label{sec:notation}
Let $x \in \R^{3\times224\times224}$ denote an input fragment image, and let
$\alpha \in \{0,1\}^{224\times224}$ denote its binary foreground mask, where
foreground pixels correspond to the visible fragment support. We consider three
ViT configurations $c \in \{\textsc{base},\textsc{mskd},\textsc{extr}\}$, where
\textsc{base} denotes the standard ViT, \textsc{mskd} the masked ViT, and
\textsc{extr} the ViT trained on inpainted fragments. For each configuration $c$,
the backbone is denoted by $V_c : \R^{3\times224\times224} \to \R^{197\times768}$;
the output sequence contains one class token and $196$ patch tokens. The
classification head is $W_c \in \R^{768\times C}$, where $C$ is the number of
target style classes. The image-level descriptor extracted from configuration $c$
using aggregation strategy $m$ is written as $g_c^{(m)} \in \R^{768}$, with
$m \in \{\cls,\mathrm{avg},\mathrm{max},\mathrm{attention}\}$. The RGB image is
divided into $196$ non-overlapping patches $\{\tilde p^{(i)}\}_{i=1}^{196}$,
$\tilde p^{(i)} \in \R^{3\times16\times16}$, and the foreground mask into $196$
mask patches $\{\bar p^{(i)}\}_{i=1}^{196}$, $\bar p^{(i)} \in \{0,1\}^{16\times16}$.
Finally, the patch embedding returned by $V_c$ for the $i$-th patch is denoted by
$t_c^{(i)} \in \R^{768}$, $i = 1,\dots,196$.
 
\begin{figure}[t]
\centering
\includegraphics[width=0.75\linewidth]{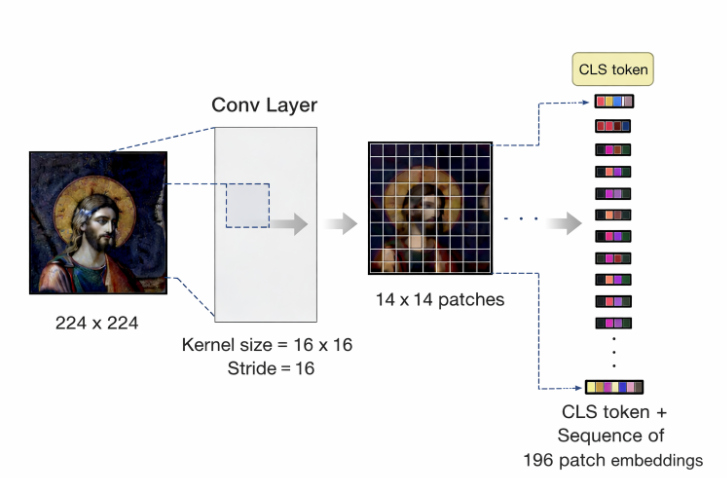}
\caption{ViT-B/16 tokenization. A $224 \times 224$ image is split into a $14 \times 14$ grid of $16 \times 16$ patches, followed by a learnable \texttt{[CLS]} token.}
\label{fig:vit_split}
\end{figure}
 
\subsection{ViT baseline}
\label{sec:baseline}
The baseline model follows the ViT-B/16 formulation of Dosovitskiy et
al.~\cite{dosovitskiy2020image}. Each fragment is resized to $224\times224$ and
passed through a patch-projection layer implemented as a convolution with kernel
size $16\times16$, stride $16$, and $768$ output channels. This produces a
$14\times14$ grid of patch embeddings, giving $196$ patch tokens in total (\cref{fig:vit_split}). A
learnable \cls{} token is appended to the sequence, and the resulting $197$ tokens
are processed by the transformer encoder. The final \cls{} representation is
passed to the classifier head $W_{\textsc{base}}$. This model does not use
fragment-specific information; it therefore provides a direct reference for
measuring the effect of foreground masking, inpainting-based regularisation, and
ensemble fusion.
 
\subsection{Masked ViT for irregular fragments}
\label{sec:masked}
Irregular fragments introduce background regions when they are placed on a square
image canvas. 
These regions do not contain painted surface, but they are still
processed as patch tokens by a standard ViT. To reduce their influence, the masked
ViT uses the foreground mask $\alpha$ to identify valid fragment tokens. The mask
$\alpha$ is split into $196$ patches using the same $16\times16$ grid as the RGB
image. For each mask patch we define a binary token indicator
\begin{equation}
  l_i = \begin{cases} 0, & \text{if } \sum_{k,q}\bar p^{(i)}_{kq} = 0,\\[2pt]
                      1, & \text{otherwise.}\end{cases}
  \label{eq:indicator}
\end{equation}

\begin{figure}[t]
\centering
\includegraphics[width=0.42\linewidth]{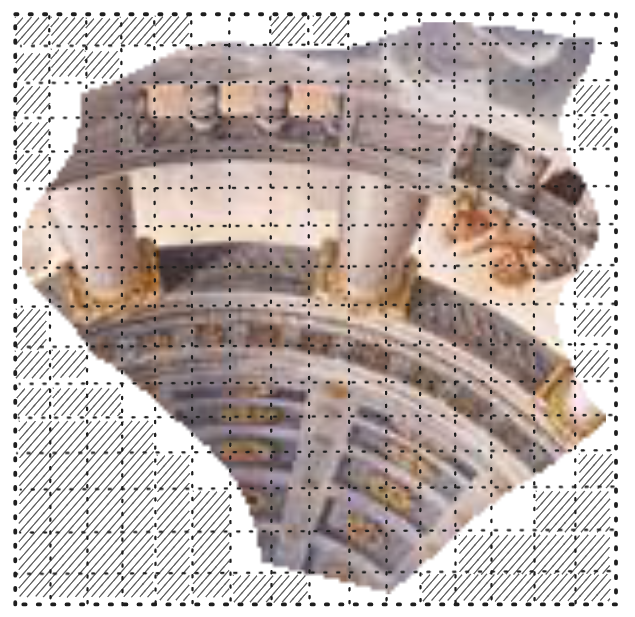}
\caption{Foreground-guided masking. Background-only tokens are suppressed.}
\label{fig:masking_process}
\end{figure}

Thus $l_i = 0$ if the patch contains only background, and $l_i = 1$ otherwise. The patch-level attention mask is then defined as $\Ahat=l\,l^\top$, with
$\Ahat_{ij}=l_i l_j$ and $l\in\{0,1\}^{196}$, so that a pair of patch tokens is
allowed to interact only when both tokens contain foreground content. The mask is extended to include the \cls{} token, indexed by $0$, with $\Ahat_{00}=1$ and $\Ahat_{0j}=\Ahat_{j0}=l_j$ for $j=1,\dots,196$.
With this construction, the \cls{} token interacts with foreground tokens but not
with background-only tokens. The same foreground information is also used during
global aggregation when pooling-based descriptors are computed. This preserves the
ViT backbone while adapting its token interactions to the visible support of the
fragment (see \cref{fig:masking_process}).

\subsection{Inpainting-based geometric regularisation}
\label{sec:inpaint}
\begin{figure}[t]
\centering
% ---------- (a) CLEOPATRA : 4 images in one PNG ----------
\begin{subfigure}[b]{\linewidth}
  \centering
  \includegraphics[width=0.95\linewidth]{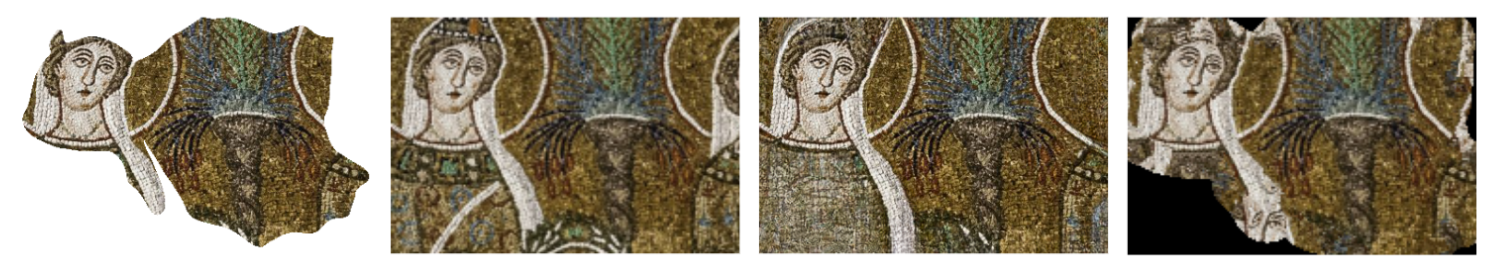}
\caption{CLEOPATRA: original fragment, ground-truth extrapolation, and inpainting by LaMa~\cite{suvorov2022resolution} and Criminisi et al.~\cite{criminisi2004region} (left to right).}

\end{subfigure}

\vspace{6pt}
% ---------- (b) POMPAAF : 3 images in one PNG ----------
\begin{subfigure}[b]{0.75\linewidth}
  \centering
  \includegraphics[width=\linewidth]{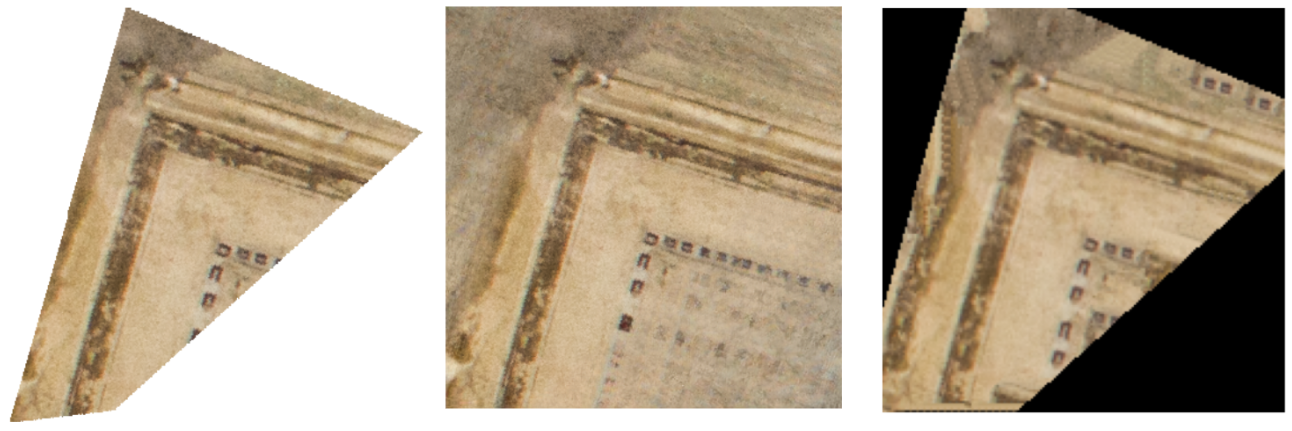}
\caption{POMPAAF: original fragment and inpainting by LaMa~\cite{suvorov2022resolution} and Criminisi et al.~\cite{criminisi2004region} (left to right).}
\end{subfigure}
\vspace{6pt}
\caption{Inpainting-based geometric regularisation. Each irregular fragment is
extrapolated to a full square so that its boundary matches the ViT patch grid;
the synthesised pixels are used only for regularisation, not as authentic evidence.
For CLEOPATRA, the removed region can be recovered exactly, providing a
ground-truth reference for the extrapolation; for POMPAAF it cannot.}
\label{fig:inpaint}
\end{figure}

Masking preserves the original fragment support. As a complementary strategy, we
also regularise fragment geometry before classification. The purpose is not to
reconstruct the original artwork, but to reduce the mismatch between irregular
fragment boundaries and the regular ViT patch grid. All pixels outside the fragment
mask are treated as target regions, while the visible fragment content is treated
as the source region. We first considered the exemplar-based inpainting formulation
of Criminisi et al.~\cite{criminisi2004region}, where missing regions are filled by
copying patches from known areas according to priority terms based on confidence
and structure. We then used LaMa inpainting~\cite{suvorov2022resolution}, which
uses Fast Fourier Convolutions to support image-wide receptive fields and
long-range spatial propagation. The inpainted images are used only as regularised
inputs for the \textsc{extr} branch (\cref{fig:inpaint}). The synthesised pixels are not treated as
authentic archaeological evidence; the classifier is trained and evaluated
empirically to determine whether this geometric regularisation improves style
recognition relative to the baseline and masked variants.

\subsection{KL-based contrastive training}
\label{sec:kl}
% CHANGE: narrative kept from Nadeem; equations corrected (purple).
The three transformer branches are trained with a classification objective. In
addition, we use a KL-based supervised contrastive term inspired by supervised
contrastive learning~\cite{khosla2020supervised}. Unlike SupCon, which
compares normalised embeddings with cosine similarity, our term compares predicted
class distributions, so that it operates in the same probabilistic space as
cross-entropy. Let $B$ denote a mini-batch and let $p_i \in \Delta^{C-1}$ be
the softmax output for sample $i$ with ground-truth label $\ell_i$.
For each anchor, we define the comparison set as $A_i=\{j\in B\mid j\neq i\}$ and the positive set as $P_i=\{j\in A_i\mid \ell_j=\ell_i\}$.
We measure agreement between two predicted distributions with a \sa{Kullback--Leibler} similarity: 
\begin{equation}
  \sa{\mathrm{KL}(p\|q) = \sum_{r} p_r \log\frac{p_r}{q_r}},
  \label{eq:skl}
\end{equation}
and define the contrastive term with the temperature inside the exponential,
so it does not cancel between numerator and denominator:
\begin{equation}
  \sa{\mathcal{L}_{\mathrm{KL}}
  = \sum_{i \in B} -\frac{1}{|P_i|} \sum_{j \in P_i}
    \log \frac{\exp\!\big(-\mathrm{KL}(p_i \| p_j)/\tau\big)}
              {\sum_{k \in A_i}\exp\!\big(-\mathrm{KL}(p_i \|p_k)/\tau\big)}}.
  \label{eq:klcon}
\end{equation}
The full training objective is applied after a short warm-up period:
\begin{equation}
  \mathcal{L} = \begin{cases}
    \mathcal{L}_{\mathrm{CE}}, & t' < t_w,\\[2pt]
    \alpha\,\mathcal{L}_{\mathrm{CE}} + \beta\,\mathcal{L}_{\mathrm{KL}}, & t' \ge t_w,
  \end{cases}
  \label{eq:total}
\end{equation}
where $t'$ is the current epoch, $t_w$ the number of warm-up epochs, $\tau$ the
temperature, and $\alpha,\beta$ control the relative contribution of the two terms.
The warm-up stage stabilises standard classification learning before introducing
distribution-level contrastive regularisation.

\subsection{Ensemble fusion}
\label{sec:fusion}
 
The three transformer branches provide complementary views of the same fragment:
the baseline branch processes the resized image directly, the masked branch
suppresses background-only tokens, and the inpainted branch processes the
regularised input.

\subsubsection{Simple learnable logit ensemble (primary).}
Our primary fusion is deliberately simple: a learnable weighted average of the
branch logits. With branch logits $z_c$ and learnable weights $w_c$, the fused
logits are $z = \sum_c w_c z_c$. The fusion module is first trained on top of
frozen branches; the branches are then gradually unfrozen and jointly optimised
with an ensemble loss that includes auxiliary branch-level classification terms,
$\mathcal{L}^{+} = \mathcal{L}(z) + \lambda\,\tfrac13\sum_c \mathcal{L}_{\mathrm{CE}}(z_c)$.
Keeping fusion at the logit level means that any difference observed in the
interpretability analysis can be attributed to branch interaction rather than to an
additional representational module.

\subsubsection{Graph-based fusion variant (secondary).}
\begin{figure}[t]
\centering
\setlength{\abovecaptionskip}{3pt}
\setlength{\belowcaptionskip}{-6pt}

\includegraphics[width=0.8\linewidth]{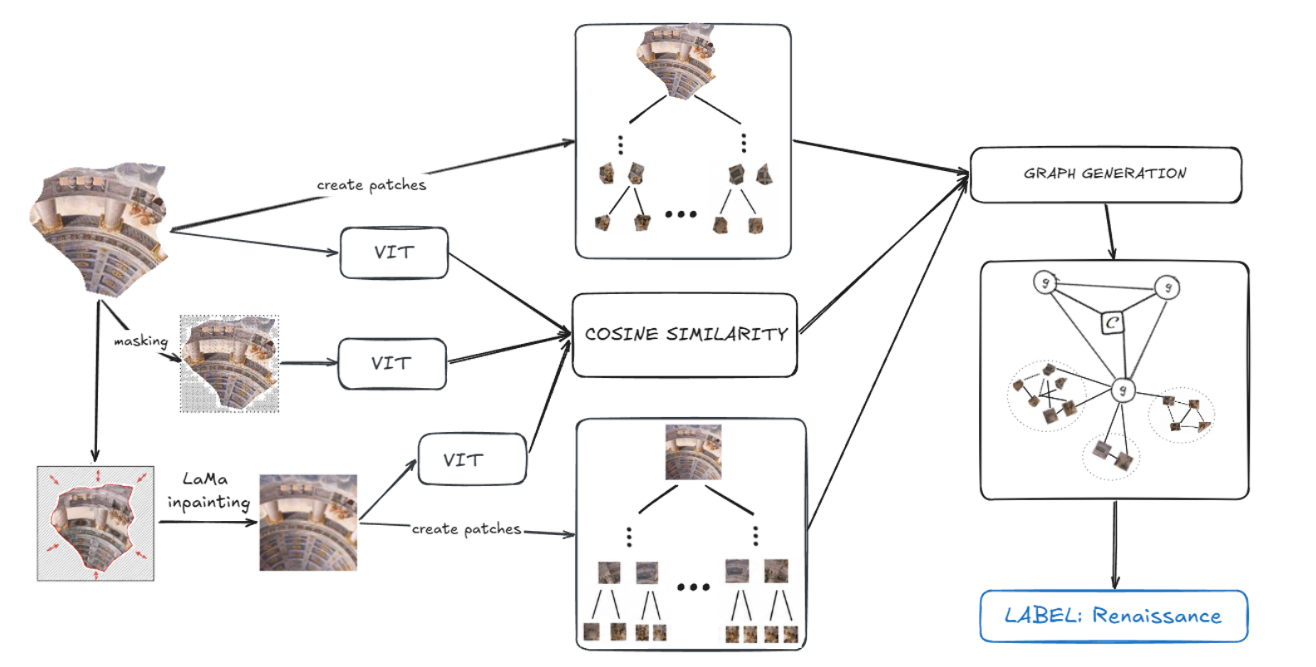}

\caption{Secondary graph-fusion variant merging patch and global representations from the three ViT branches for final prediction.}
\label{fig:graph}

\end{figure}

% CHANGE: Nadeem's graph method kept but condensed; framed as secondary (purple).
As a secondary comparison, we also evaluated a graph-based fusion variant.
For each branch $c$ we construct a graph whose nodes are the patch embeddings
$\{t_c^{(i)}\}_{i=1}^{196}$. Patch connectivity is derived from a
Binary-Partition-Tree (BPT) inspired hierarchy~\cite{rashid2026shapbpt}: adjacent
patches are merged bottom-up into progressively larger coalitions, candidate
patch-to-patch edges are read from the hierarchy, and only edges whose patch
embeddings exceed a cosine-similarity threshold are retained. Each branch graph is
augmented with a global node $g_c^{(m)}$ connected to compatible patch nodes. The
three branch graphs are then merged into a single ensemble graph, guided by the
predicted class probabilities of the branches, and a central virtual node is
connected to the three global nodes. The merged graph is processed by a graph
attention network, and the prediction is read from the central node (\cref{fig:graph}). We report this variant only as a comparison: as shown in
\cref{sec:res-cleo,sec:res-pomp}, its gain over the simple logit ensemble is
small and inconsistent relative to its cost, so we do not adopt it as the main
model and use its patch-relation graph only as an additional interpretive view.

%We report this variant only as a comparison: as shown in \cref{sec:res-cleo,sec:res-pomp}, it does not consistently outperform the simple logit ensemble, so we do not adopt it as the main model and use its patch-relation graph only as an additional interpretive view.

\subsection{Post-hoc interpretation}
\label{sec:interp-methods}
% Nadeem's rollout + SHAP-BPT kept (black); controlled-ablation method added
% (purple) -- this analysis was already performed, so no new experiment is needed.
For analysis we use post-hoc explanation methods together with a controlled
input-ablation study. Attention Rollout~\cite{abnar2020quantifying} propagates
attention weights across transformer layers to estimate token-level relevance.
Given attention matrices $A^{(l)}$ at layer $l$, residual connections are incorporated as $\tilde A^{(l)}=\alpha A^{(l)}+(1-\alpha)I$, and the cumulative rollout is computed recursively as
$R^{(l)} = \tilde A^{(l-1)} R^{(l-1)}$ with $R^{(0)} = I$, producing a patch-level
relevance map over the ViT input tokens. We also use
SHAP-BPT~\cite{rashid2026shapbpt}, which assigns Shapley-style attributions over
data-aware image partitions built by merging adjacent regions according to a
distance based on colour, area, and perimeter; the resulting hierarchy provides
region-level explanations aligned with image morphology rather than isolated pixel
attributions. \sa{From these region-level attributions we also derive % a single quantitative summary, the \textit{foreground-relevance ratio} (\textit{FRR}), which is the fraction of positive attribution that falls inside original fragment mask (so that any attribution on the LaMa-synthesised border counts as outside); we report it in 
a region-level foreground-relevance analysis: projecting the original fragment mask onto each map, we measure the fraction of positive attribution in the fragment interior, the fragment boundary, and the background, where the background of the inpainted branch is the LaMa-synthesised surround (Sec. \ref{sec:res-interp}).}
In addition to these attribution methods, we run a controlled input-ablation
study that perturbs one factor at a time and measures the resulting change in the
model output (the drop in correct-class probability, the prediction-flip rate, and
changes in softmax margin and entropy). %The perturbations target colour(grayscale, desaturation, hue, mean-colour replacement), texture and detail (blur,low resolution, high-pass filtering, noise, JPEG compression, pixel/patch shuffle),structure (edge-only and alpha-mask-only inputs), frequency content(low/high/mid-pass and random-phase), individual colour channels, and centreversus border regions. 
\sa{The perturbations span colour, texture and detail, structure (including edge-only and alpha-mask-only inputs), frequency content, individual colour channels, and centre-versus-border regions.} Unlike attention maps, this provides a functional estimate
of which visual information each model actually depends on, and in particular
whether decisions rely on internal painted content or merely on the fragment
silhouette.
 
% =====================================================================
\section{Experiments and Results}
\label{sec:exp}
 
\subsection{Datasets}
\label{sec:data}
We evaluate the framework on two fresco-fragment style-classification benchmarks,
CLEOPATRA and POMPAAF. All experiments are at fragment level: each fragment is an
independent sample, and the model is given no information about the source artwork.
CLEOPATRA~\cite{cascone2023classification} contains eleven artistic-style classes
and provides complete artworks together with their fragmented representations. The
split is defined at artwork level, so that all fragments of a painting remain in
the same partition: $26$ training, $6$ validation, and $8$ test artworks,
corresponding to $18{,}458$ training, $4{,}286$ validation, and $5{,}713$ test
fragments. Because the split is at artwork level, the eight test artworks over
eleven classes make absolute numbers sensitive to the particular split, which we
keep in mind when interpreting results. POMPAAF~\cite{elkin2025recognizing} is a
Pompeian wall-painting fragment dataset organised by four styles (structural,
architectural, ornamental, intricate), built from $311$ full frescoes under several
fragmentation protocols. We evaluate the settings for which explicit style
annotations are available: crossing cuts with $5$ and $20$ cuts, square fragments
with $12$ and $160$ fragments, and eroded Voronoi fragments with $12$ and $160$
fragments; the convex setting is excluded because it lacks the required style
annotations. \Cref{tab:data} summarises the evaluation sets.
 
\begin{table}[!t] \centering  \scriptsize \setlength{\tabcolsep}{3pt} \renewcommand{\arraystretch}{0.82} \caption{Fragment-level evaluation sets.} \label{tab:data} \begin{tabular}{@{}lccc@{}} \toprule Dataset / setting & Classes & Train / Val & Test \\ \midrule CLEOPATRA & 11 & 18{,}458 / 4{,}286 & 5{,}713 \\ POMPAAF cross-5 & 4 & 3{,}160 / -- & 837 \\ POMPAAF cross-20 & 4 & 37{,}340 / -- & 9{,}542 \\ POMPAAF square-12 & 4 & 2{,}964 / -- & 768 \\ POMPAAF square-160 & 4 & 38{,}733 / -- & 10{,}057 \\ POMPAAF eroded-12 & 4 & 2{,}964 / -- & 768 \\ POMPAAF eroded-160 & 4 & 39{,}600 / -- & 10{,}240 \\ \bottomrule \end{tabular} \end{table}
 
\subsection{Evaluation protocol}
\label{sec:eval}
We report accuracy and macro F1-score. Accuracy measures the overall proportion of
correctly classified fragments, while macro F1-score gives equal weight to each
class and is therefore more informative when the class distribution is uneven. For
CLEOPATRA, hyperparameters are selected on the validation split; after model
selection, the final model is trained on the union of the training and validation
fragments and evaluated on the held-out test split. For POMPAAF, we follow the
predefined train/test partitions provided for each fragmentation setting; as no
validation split is provided, model selection relies on training-set cross-validation only. \sa{For both datasets we use our own reproduction of Elkin et al. \cite{elkin2025recognizing} as the baseline (Table~\ref{tab:cleo}), obtained by running their publicly released code under their published protocol. On CLEOPATRA our reproduction is close to the originally reported scores (e.g. 0.407 vs. 0.470 accuracy), whereas on POMPAAF we were unable to reproduce the originally reported numbers under the same data splits, fragmentation settings, and training strategy. We therefore report reproduced numbers throughout, so that all methods are compared under identical, verifiable conditions.}
%\sa{Importantly, the Elkin et al. \cite{elkin2025recognizing} baseline is not taken from their paper; instead, running their publicly released code under their own data splits, fragmentation settings, and training strategy, we were unable to reproduce the originally reported scores in our experiments. We therefore use our own reproduction as the baseline (Table~\ref{tab:cleo}), ensuring that all methods are compared under identical, verifiable conditions.} %For POMPAAF, we reproduced the comparison results ourselves by running the original code under the same data partitions, fragmentation settings, and evaluation strategy as Elkin et al. \cite{elkin2025recognizing}.
 
\subsection{Training protocol}
\label{sec:protocol}
All transformer-based models use an input resolution of $224\times224$. We compare
three single-branch models, namely, the standard ViT-B/16 baseline, the masked ViT, and
the ViT trained on inpainted fragments, and combine them at the fusion stage.
Training follows a progressive fine-tuning strategy: the pretrained backbone is
first frozen and only the task-specific classifier is optimised, and the backbone
is then gradually unfrozen as validation improvement saturates. Weight decay is
applied to trainable parameters except bias and LayerNorm terms, and optimisation
uses Adam with cosine learning-rate scheduling. For the ensemble, the same
principle is applied: the fusion module is first trained on top of frozen branch
representations, after which the branches are gradually unfrozen and jointly
optimised with the auxiliary branch-level classification terms.
% TODO (optional): add exact batch size, base LR, epochs, t_w, tau, alpha/beta, lambda, hardware.
 
\subsection{Results on CLEOPATRA}
\label{sec:res-cleo}
\Cref{tab:cleo} reports the results on CLEOPATRA. \sa{For reference, the original CLEOPATRA study of Cascone et al. \cite{cascone2023classification} reports 0.28 accuracy and 0.27 macro-F1 with a random-forest classifier, and our reproduction of Elkin et al. \cite{elkin2025recognizing} reaches 0.407/0.403; the ViT-based models improve substantially over both prior baselines.} The standard ViT-B/16 baseline
reaches $0.604$ accuracy and $0.596$ macro F1-score with cross-entropy training.
The KL-based contrastive objective gives a small improvement for all three
single-branch models. The masked ViT improves over the baseline, indicating that
suppressing background-only tokens is useful for irregular fragments, and the
inpainted branch gives the strongest single-model result, reaching $0.636$ accuracy
and $0.629$ macro F1-score. The simple learnable logit ensemble performs best
among the models we adopt, reaching $0.656$ accuracy and $0.648$ macro F1-score,
an improvement of $5.2$ and $5.2$ points over the ViT-B/16 baseline. The graph-fusion variant reaches $0.681$/$0.677$, slightly above the simple ensemble; this small gain, however, is dataset-specific, as the two are effectively tied on POMPAAF (differences $\leq 0.004$; \cref{sec:res-pomp}). Since graph fusion is not inferior but offers only a small, inconsistent improvement at a substantial architectural and computational cost, we adopt the simple ensemble as our main model on grounds of parsimony and report graph fusion as a secondary comparison.
%; however, as shown on POMPAAF(\cref{sec:res-pomp}) this advantage does not carry over to the other benchmark,so we report it as a secondary comparison and treat the simple ensemble as our mainmodel.
% TODO (no new experiment required): if a clean re-run of the simple ensemble on
% CLEOPATRA is available, reconcile the 0.656 (long version) vs 0.681 (graph) gap
% and, ideally, add mean +/- std over seeds. Not blocking for submission.
 
\begin{table}[!t] \centering \scriptsize \setlength{\tabcolsep}{3pt} \renewcommand{\arraystretch}{0.82} \caption{CLEOPATRA results. CE: cross-entropy; KL: KL-based contrastive objective.} \label{tab:cleo} \begin{tabular}{@{}lcccc@{}} \toprule Model & Acc. CE & F1 CE & Acc. KL & F1 KL \\ \midrule 
%Elkin et al.~\cite{elkin2025recognizing}\sa{(reproduced)} & 0.475 & 0.466 & -- & -- 
Cascone et al. \cite{cascone2023classification} & 0.28 & 0.27 & -- & -- \\
Elkin et al.~\cite{elkin2025recognizing}\sa{(reproduced)} & 0.407 & 0.403 & -- & -- 
\\ ViT-B/16 baseline & 0.604 & 0.596 & 0.611 & 0.599 \\ Masked ViT & 0.616 & 0.601 & 0.624 & 0.614 \\ Inpainted ViT & 0.623 & 0.612 & 0.636 & 0.629 \\ \midrule Logit ensemble & -- & -- & \textbf{0.656} & \textbf{0.648} \\ Graph fusion & -- & -- & 0.681 & 0.677 \\ \bottomrule \end{tabular} \end{table}
 
\subsection{Results on POMPAAF}
\label{sec:res-pomp}
\Cref{tab:pomp} summarises the POMPAAF results. For each fragmentation setting we
compare the best single ViT branch with the simple ensemble, and report the
graph-fusion variant alongside for completeness. The ensemble improves over the
best single branch in four of the six settings, with the strongest gains for square
fragments with $12$ fragments and eroded fragments with $12$ fragments. In the two
most difficult settings, i.e., crossing cuts with $20$ cuts and eroded fragments with
$160$ fragments, the best single ViT remains slightly better, although the
ensemble stays close in both accuracy and macro F1-score. Crucially, the
graph-fusion variant matches the simple ensemble in every setting (differences
$\le 0.004$), which indicates that the complementary signal is already captured at
the logit level and that the added architectural and computational complexity of graph fusion brings no measurable benefit at this fragmentation level. %the additional graph machinery is not warranted. 
These
results indicate that fusion is most useful when fragments retain enough local
structure for the branches to provide complementary evidence; under stronger
fragmentation the available stylistic evidence becomes weaker and fusion does not
always improve over the strongest single branch.
 
\begin{table}[!t] \centering  \scriptsize \setlength{\tabcolsep}{3pt} \renewcommand{\arraystretch}{0.82} \caption{POMPAAF results, reported as accuracy / macro F1.} \label{tab:pomp} \begin{tabular}{@{}lccc@{}} \toprule Setting & Best ViT & Ensemble & Graph \\ \midrule Cross-5 & 0.611 / 0.613 & \textbf{0.624 / 0.632} & 0.625 / 0.632 \\ Cross-20 & \textbf{0.525 / 0.526} & 0.516 / 0.523 & 0.516 / 0.523 \\ Square-12 & 0.683 / 0.690 & \textbf{0.714 / 0.718} & 0.715 / 0.718 \\ Square-160 & 0.634 / 0.643 & \textbf{0.638 / 0.643} & 0.639 / 0.644 \\ Eroded-12 & 0.669 / 0.670 & \textbf{0.694 / 0.691} & 0.694 / 0.691 \\ Eroded-160 & \textbf{0.606 / 0.613} & 0.600 / 0.611 & 0.600 / 0.611 \\ \bottomrule \end{tabular} \end{table}
 
\subsection{Interpretability results}
\label{sec:res-interp}
\begin{figure}[t]
\centering
\setlength{\abovecaptionskip}{3pt}
\setlength{\belowcaptionskip}{-6pt}

% (a) Ablation row
\begin{minipage}{0.95\linewidth}
\centering
\includegraphics[width=0.82\linewidth]{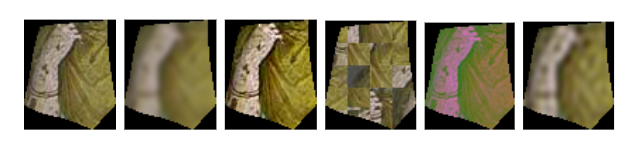}\\[-2pt]
{\small \textbf{(a)} Controlled ablation transforms.}
\end{minipage}

%\vspace{5pt}

% (b) Baseline
\begin{minipage}{0.95\linewidth}
\centering
\includegraphics[width=\linewidth]{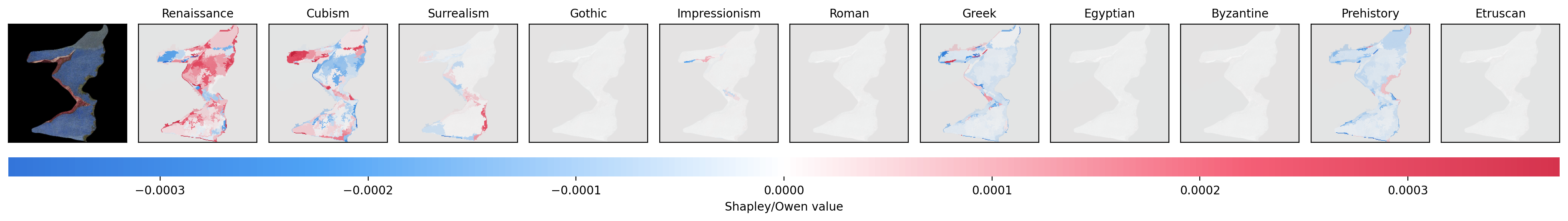}\\[-2pt]
{\small \textbf{(b)} Baseline ViT.}
\end{minipage}

%\vspace{4pt}

% (c) Masked
\begin{minipage}{0.95\linewidth}
\centering
\includegraphics[width=\linewidth]{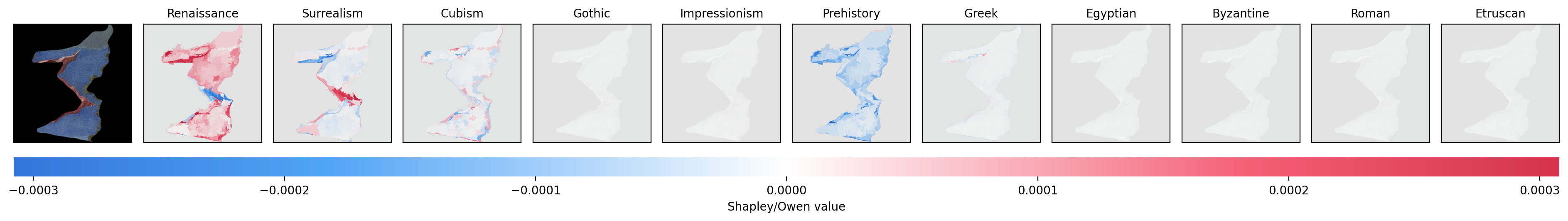}\\[-2pt]
{\small \textbf{(c)} Masked ViT.}
\end{minipage}

%\vspace{4pt}

% (d) Inpainted
\begin{minipage}{0.95\linewidth}
\centering
\includegraphics[width=\linewidth]{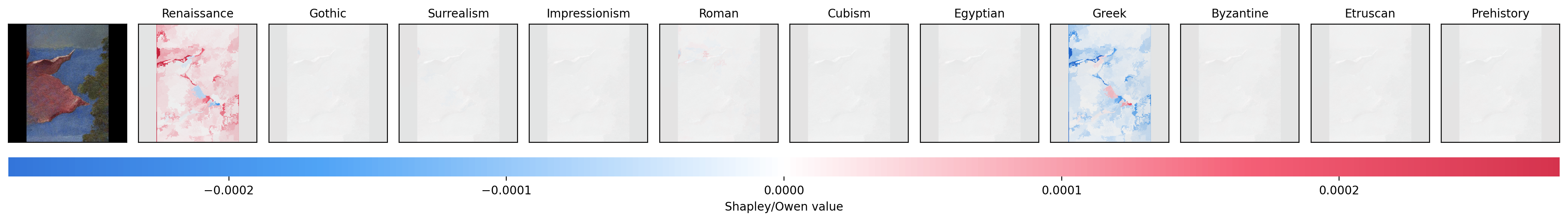}\\[-2pt]
{\small \textbf{(d)} Inpainted ViT.}
\end{minipage}

\caption{\sa{Interpretability analysis on a CLEOPATRA fragment. (a) Controlled ablation transforms. (b)–(d) SHAP-BPT attribution maps for the baseline, masked, and inpainted branches across style classes. The baseline and masked branches concentrate attribution on the painted fragment, whereas the inpainted branch, although still interior-dominant, assigns a larger share to the synthesised surround (quantified in Table \ref{tab:region}).}
%Interpretability analysis on a CLEOPATRA fragment. (a) Controlled ablation transforms. (b)--(d) SHAP-BPT attribution maps for the baseline, masked, and inpainted branches. The baseline and masked branches focus mainly on painted fragment content, whereas the inpainted branch places stronger attribution near the synthesised outer border.
}
\label{fig:interp}

\end{figure}

% CHANGE: NEW subsection. Reports the QUALITATIVE findings already produced in the
% extended analysis. No invented numeric tables. Reuse the existing figures.
We analyse the baseline, masked, inpainted, and ensemble models with attention
rollout, SHAP-BPT, and the controlled input-ablation study, on both POMPAAF and
CLEOPATRA. %The conclusions below are qualitative summaries of that analysis;
The conclusions below combine a quantitative region-level foreground-relevance summary (Table \ref{tab:region}) with qualitative attribution and ablation analysis; representative attribution maps and example ablation transforms are shown in
\cref{fig:interp}.

\paragraph{\textbf{Region-level attribution.}} \sa{To localise the evidence each branch uses, we project the original fragment mask onto every SHAP-BPT map and measure the fraction of positive attribution in the fragment interior, the fragment boundary, and the background (Table \ref{tab:region}). All three branches place the majority of their attribution on the real fragment (interior plus boundary $\ge$ 0.72), confirming that they rely on genuine painted content rather than on the empty or synthesised surround. The branches differ mainly in the background: the baseline and masked branches assign it only $0.096$ and $0.106$ of their attribution, whereas the inpainted branch assigns it $0.284$ (roughly three times as much). This gap is large relative to the baseline and masked variability (std 0.069 and 0.077), so it reflects a consistent tendency rather than a few outliers, although the inpainted branch itself shows high per-fragment variance (std 0.161). Inpainting therefore improves geometric regularisation, i.e., it yields the strongest single branch (Table \ref{tab:cleo}), but at an interpretability cost as a measurable part of its evidence comes from non-authentic, synthesised pixels, which is an important caution for heritage use.}

\sa{A per-image comparison %is more informative than the means. The 
reinforces this difference: the 
baseline and masked branches are strongly correlated in their per-fragment foreground reliance ($r = 0.74$), reflecting near-identical attention to the painted interior, whereas the inpainted branch correlates only weakly with the other two ($r = 0.47$ and $0.50$), %Thus, although its average reliance is comparable, the inpainted branch attends to different regions on a substantial fraction of fragments, consistent with the occasional border-oriented attribution illustrated in Fig. \ref{fig:interp}. We therefore characterise the inpainted branch's reliance on synthesised content as a per-example behaviour rather than a systematic shortcut.
confirming that it distributes its attention differently, in line with its higher background share in Table \ref{tab:region}.}

\begin{table}[t]
\centering
\caption{Region-level SHAP-BPT attribution on CLEOPATRA, reported as positive-attribution fractions (mean $\pm$ std, 100 test fragments).}
\label{tab:region}
\begin{tabular}{lccc}
\toprule
Branch & Fragment interior & Fragment boundary & Background \\
\midrule
Baseline ViT  & $0.561 \pm 0.160$ & $0.343 \pm 0.128$ & $0.096 \pm 0.069$ \\
Masked ViT    & $0.570 \pm 0.156$ & $0.324 \pm 0.114$ & $0.106 \pm 0.077$ \\
Inpainted ViT & $0.513 \pm 0.173$ & $0.204 \pm 0.048$ & $\mathbf{0.284 \pm 0.161}$ \\
\bottomrule
\end{tabular}
\end{table}

\paragraph{\textbf{Models rely on internal structure, not the silhouette.}}
Across both datasets, the perturbations that most degrade prediction stability and
confidence are those that remove or distort information \emph{inside} the fragment:
replacing the interior with its mean colour, strong blur, pixel shuffling, and
random-phase (frequency) perturbations. In particular, keeping only the alpha mask
(the silhouette) causes a large drop, showing that fragment shape alone is not
sufficient and that the decision is driven by internal texture, edge structure,
and frequency content. This holds for the baseline and masked branches and is
consistent with the motivation behind masking: restricting the model to the valid
region does not make it rely on the binary shape but on the painted content within
it. 

\paragraph{\textbf{SHAP-BPT confirms a fragment-centred focus.}}
For the baseline and masked branches, attribution concentrates within the fragment
region; whether it is more edge-dominated or interior-dominated depends on the
fragmentation type (more edge-dominated for crossing cuts, more interior-focused
for square fragments and for CLEOPATRA). %The inpainted branch behaves differently: its dominant attribution is associated with a broad outer-border region rather than with the fragment interior, i.e.\ it partly grounds its decision on synthesised content. This is an important caution for heritage use as inpainting regularises geometry, but the resulting model can rely on non-authentic pixels.
%\sa{In individual examples such as the inpainted panel of Fig. \ref{fig:interp}, the inpainted branch can place stronger attribution on a broad outer-border region than on the fragment interior, i.e. in those cases it partly grounds its decision on synthesised content. Averaged over the test fragments, however, its foreground-relevance is comparable to the other branches (Table \ref{tab:frr}), so we treat this border attribution as a per-example caution rather than a systematic effect. It nonetheless matters for heritage use, since inpainting regularises geometry but the resulting model can, in such cases, rely on non-authentic pixels.} Encouragingly,the ensemble does not inherit this behaviour, i.e., its 
\sa{The inpainted branch remains interior-dominant but, consistent with Table \ref{tab:region}, distributes a visible share of its attribution onto the synthesised surround (Fig. \sa{\ref{fig:interp}}). The ensemble's} attribution remains
fragment-centred while being slightly more spatially distributed, consistent with
combining a locally sensitive branch with a more global one.

\label{sec:summary}
Across both datasets the results show consistent trends. First, the standard ViT baseline is a competitive reference, but it treats foreground and background tokens uniformly. Second, masking and inpainting address different aspects of the fragment
problem: masking suppresses background-only tokens while preserving the true
support, whereas inpainting regularises the input geometry at the cost of
introducing synthesised content the model may rely on. Third, a simple %learnable logit ensemble captures most of the complementary signal across branches, and the more complex graph variant does not consistently improve on it. 
learnable logit ensemble captures most of the complementary signal across branches, and the more complex graph variant adds only a small, inconsistent gain that does not justify its complexity. Finally, the interpretability analysis shows that the models rely on internal
painted evidence rather than on silhouette or background; \sa{a region-level analysis additionally shows that the inpainting-based branch draws roughly three times as much attribution from the synthesised surround as the other branches (Table \ref{tab:region}), a quantified caution for heritage use.}
Overall, the results support the central hypothesis that fresco-fragment style classification benefits from explicitly modelling irregular support and local visual evidence. 
% =====================================================================
\section{Discussion}
\label{sec:discussion}
% Nadeem's discussion kept; graph paragraph reframed (purple); inpainting-shortcut
% and limitations added (purple).
%The results show that fresco-fragment style classification benefits from treating fragmentation as a specific visual condition rather than a standard image-classification setting. A plain ViT-B/16 provides a strong baseline, but it processes all patch tokens uniformly, including background regions introduced by irregular fragment geometry. This is not ideal for archaeological fragments, where the stylistic evidence is carried by the painted surface and not by the surrounding empty canvas. The masked ViT addresses this directly by suppressing background-only tokens; its improvement over the baseline indicates that foreground-aware token selection is useful when the valid support occupies only part of the input. 

{Fresco-fragment style classification should be treated as a fragment-specific problem: a plain ViT-B/16 is strong, but it processes background introduced by irregular geometry as ordinary tokens. Masked ViT addresses this by suppressing background-only tokens, and its gain over the baseline supports foreground-aware token selection.}

%The inpainted branch provides a complementary form of adaptation by regularising the input geometry; its performance suggests that reducing the mismatch between fragment shape and the regular patch grid can help, provided---as our interpretability analysis stresses---that the synthesised regions are treated only as preprocessing support and not as authentic evidence, since the inpainted branch partly attributes its decisions to that synthesised border content.
\sa{The inpainted branch complements this adaptation by regularising input geometry, suggesting that reducing the mismatch between fragment shape and the regular patch grid can help, provided that synthesised regions are treated only as preprocessing support rather than authentic evidence; our region-level analysis (Sec. \ref{sec:res-interp}) shows that the inpainted branch, while remaining interior-dominant, draws about three times as much attribution from the synthesised surround as the baseline and masked branches.}
%On fusion, the simple learnable logit ensemble gives the strongest result among the models we adopt, improving the ViT-B/16 baseline from $0.604$ to $0.656$ accuracy and from $0.596$ to $0.648$ macro F1-score on CLEOPATRA. A more complex graph-fusion variant reaches a higher CLEOPATRA number but matches the simple ensemble on POMPAAF, so the gain does not generalise and we do not attribute it to the graph structure. We therefore present fusion as a way to combine complementary branch evidence rather than as a central contribution, and use the interpretability analysis to characterise how the individual and fused models differ. On POMPAAF, the benefit of fusion depends on the fragmentation regime: it helps most when the fragments retain enough local structure for the branches to remain complementary, and helps less under the most severe fragmentation. \newline
{The simple logit ensemble improves the adopted CLEOPATRA baseline from $0.604/0.596$ to $0.656/0.648$ accuracy/macro-F1. Although graph fusion is higher on CLEOPATRA, it matches the simple ensemble on POMPAAF, so we treat it as a secondary comparison rather than evidence that graph structure is necessary. Fusion helps most when fragments retain enough local structure and less under severe fragmentation.}
\newline
\textbf{\textit{Limitations.}} Classification is performed at fragment level: each fragment is treated independently, without using its original artwork or its spatial relation to other fragments, so the broader archaeological context available in reassembly tasks is not exploited. CLEOPATRA's artwork-level split leaves only eight test artworks, which inflates the variance of absolute numbers. Inpainting-based regularisation must be interpreted cautiously, since the synthesised regions are not archaeological evidence and \sa{attract a measurable share of the inpainted branch's attribution (Table~\ref{tab:region})}. Finally, the reported improvements are from single runs; \sa{multi-seed evaluation would be needed to establish statistical significance.} 
% =====================================================================
\section{Conclusion}
\label{sec:conclusion}
% Nadeem's conclusion kept; headline number changed to the simple ensemble and an
% interpretability sentence added (purple).
This work studied artistic style classification from fragmented fresco images using
a progressive transformer-based framework. Starting from a standard ViT-B/16
baseline, we introduced two fragment-aware strategies. \sa{The first is} foreground-guided masking
to reduce the influence of background tokens, and \sa{the second is} inpainting-based geometric
regularisation to reduce the mismatch between irregular fragment supports and the regular ViT patch grid. \sa{These strategies are combined with} a \sa{KL} contrastive
objective and a simple learnable logit ensemble of the branches. The experiments
show that fragment-aware modelling improves style recognition from incomplete
fresco fragments. On CLEOPATRA, the simple ensemble achieves the best result
among the models we adopt, reaching $0.656$ accuracy and $0.648$ macro F1-score, while a more complex graph-fusion variant offers only a small, dataset-specific
gain and is reported as a secondary comparison. %while a more complex graph-fusion variant does not consistently help and is reported as a secondary comparison. 
On POMPAAF, fusion improves over the best
single ViT in most settings and remains competitive under stronger fragmentation.
A post-hoc interpretability analysis further shows that the models rely on
internal stylistic evidence rather than on silhouette or background, \sa{and quantifies that inpainting-based regularisation draws part of the model's evidence (about 28\% of its attribution) from synthesised pixels, i.e., a caution for heritage use.} %, and reveals that the inpainted branch partly grounds its decisions on synthesised content.
These results support the central finding that stylistic information can remain
locally recoverable under fragmentation, but is better exploited when the model
explicitly accounts for irregular support and local visual evidence. Future
work will couple fragment-level prediction with reassembly constraints and
uncertainty estimation, and will add multi-seed evaluation to confirm the observed
improvements.
 
% =====================================================================
% \section*{Acknowledgements}  % omit for double-blind submission
 
\bibliographystyle{splncs04}
\bibliography{ref}

@article{criminisi2004region,
  title={Region filling and object removal by exemplar-based image inpainting},
  author={Criminisi, Antonio and P{\'e}rez, Patrick and Toyama, Kentaro},
  journal={IEEE Transactions on image processing},
  volume={13},
  number={9},
  pages={1200--1212},
  year={2004},
  publisher={IEEE}
}

@article{dosovitskiy2020image,
  title={An image is worth 16x16 words: Transformers for image recognition at scale},
  author={Dosovitskiy, Alexey},
  journal={arXiv preprint arXiv:2010.11929},
  year={2020}
}

@inproceedings{touvron2021training,
  title={Training data-efficient image transformers \& distillation through attention},
  author={Touvron, Hugo and Cord, Matthieu and Douze, Matthijs and Massa, Francisco and Sablayrolles, Alexandre and J{\'e}gou, Herv{\'e}},
  booktitle={International conference on machine learning},
  pages={10347--10357},
  year={2021},
  organization={PMLR}
}

@inproceedings{liu2021swin,
  title={Swin transformer: Hierarchical vision transformer using shifted windows},
  author={Liu, Ze and Lin, Yutong and Cao, Yue and Hu, Han and Wei, Yixuan and Zhang, Zheng and Lin, Stephen and Guo, Baining},
  booktitle={Proceedings of the IEEE/CVF international conference on computer vision},
  pages={10012--10022},
  year={2021}
}

@inproceedings{touvron2021going,
  title={Going deeper with image transformers},
  author={Touvron, Hugo and Cord, Matthieu and Sablayrolles, Alexandre and Synnaeve, Gabriel and J{\'e}gou, Herv{\'e}},
  booktitle={Proceedings of the IEEE/CVF international conference on computer vision},
  pages={32--42},
  year={2021}
}

@inproceedings{chen2021crossvit,
  title={Crossvit: Cross-attention multi-scale vision transformer for image classification},
  author={Chen, Chun-Fu Richard and Fan, Quanfu and Panda, Rameswar},
  booktitle={Proceedings of the IEEE/CVF international conference on computer vision},
  pages={357--366},
  year={2021}
}

@article{liang2022not,
  title={Not all patches are what you need: Expediting vision transformers via token reorganizations},
  author={Liang, Youwei and Ge, Chongjian and Tong, Zhan and Song, Yibing and Wang, Jue and Xie, Pengtao},
  journal={arXiv preprint arXiv:2202.07800},
  year={2022}
}

@inproceedings{cheng2024putting,
  title={Putting the object back into video object segmentation},
  author={Cheng, Ho Kei and Oh, Seoung Wug and Price, Brian and Lee, Joon-Young and Schwing, Alexander},
  booktitle={Proceedings of the IEEE/CVF Conference on Computer Vision and Pattern Recognition},
  pages={3151--3161},
  year={2024}
}

@article{tsesmelis2024re,
  title={Re-assembling the past: The RePAIR dataset and benchmark for real world 2D and 3D puzzle solving},
  author={Tsesmelis, Theodore and Palmieri, Luca and Khoroshiltseva, Marina and Islam, Adeela and Elkin, Gur and Shahar, Ofir I and Scarpellini, Gianluca and Fiorini, Stefano and Ohayon, Yaniv and Alali, Nadav and others},
  journal={Advances in Neural Information Processing Systems},
  volume={37},
  pages={30076--30105},
  year={2024}
}

@inproceedings{elkin2025recognizing,
  title={Recognizing artistic style of archaeological image fragments using deep style extrapolation},
  author={Elkin, Gur and Shahar, Ofir Itzhak and Ohayon, Yaniv and Alali, Nadav and Ben-Shahar, Ohad},
  booktitle={International Conference on Human-Computer Interaction},
  pages={115--131},
  year={2025},
  organization={Springer}
}

@article{ferrara2025cow,
  title={The Cow of Rembrandt-Analyzing Artistic Prompt Interpretation in Text-to-Image Models},
  author={Ferrara, Alfio and Picascia, Sergio and Rocchetti, Elisabetta},
  journal={arXiv preprint arXiv:2507.23313},
  year={2025}
}

@article{van2015toward,
  title={Toward Discovery of the Artist's Style: Learning to recognize artists by their artworks},
  author={Van Noord, Nanne and Hendriks, Ella and Postma, Eric},
  journal={IEEE Signal Processing Magazine},
  volume={32},
  number={4},
  pages={46--54},
  year={2015},
  publisher={IEEE}
}

@inproceedings{ribeiro2016should,
  title={" Why should i trust you?" Explaining the predictions of any classifier},
  author={Ribeiro, Marco Tulio and Singh, Sameer and Guestrin, Carlos},
  booktitle={Proceedings of the 22nd ACM SIGKDD international conference on knowledge discovery and data mining},
  pages={1135--1144},
  year={2016}
}

@article{lundberg2017unified,
  title={A unified approach to interpreting model predictions},
  author={Lundberg, Scott M and Lee, Su-In},
  journal={Advances in neural information processing systems},
  volume={30},
  year={2017}
}

@inproceedings{selvaraju2017grad,
  title={Grad-cam: Visual explanations from deep networks via gradient-based localization},
  author={Selvaraju, Ramprasaath R and Cogswell, Michael and Das, Abhishek and Vedantam, Ramakrishna and Parikh, Devi and Batra, Dhruv},
  booktitle={Proceedings of the IEEE international conference on computer vision},
  pages={618--626},
  year={2017}
}

@article{ying2019gnnexplainer,
  title={Gnnexplainer: Generating explanations for graph neural networks},
  author={Ying, Zhitao and Bourgeois, Dylan and You, Jiaxuan and Zitnik, Marinka and Leskovec, Jure},
  journal={Advances in neural information processing systems},
  volume={32},
  year={2019}
}

@article{khosla2020supervised,
  title={Supervised contrastive learning},
  author={Khosla, Prannay and Teterwak, Piotr and Wang, Chen and Sarna, Aaron and Tian, Yonglong and Isola, Phillip and Maschinot, Aaron and Liu, Ce and Krishnan, Dilip},
  journal={Advances in neural information processing systems},
  volume={33},
  pages={18661--18673},
  year={2020}
}

@article{abnar2020quantifying,
  title={Quantifying attention flow in transformers},
  author={Abnar, Samira and Zuidema, Willem},
  journal={arXiv preprint arXiv:2005.00928},
  year={2020}
}

@article{luo2020parameterized,
  title={Parameterized explainer for graph neural network},
  author={Luo, Dongsheng and Cheng, Wei and Xu, Dongkuan and Yu, Wenchao and Zong, Bo and Chen, Haifeng and Zhang, Xiang},
  journal={Advances in neural information processing systems},
  volume={33},
  pages={19620--19631},
  year={2020}
}

@inproceedings{chefer2021transformer,
  title={Transformer interpretability beyond attention visualization},
  author={Chefer, Hila and Gur, Shir and Wolf, Lior},
  booktitle={Proceedings of the IEEE/CVF conference on computer vision and pattern recognition},
  pages={782--791},
  year={2021}
}

@article{tran2025novel,
  title={A novel approach based on graph attention networks for fruit recognition},
  author={Tran-Anh, Dat and Vu, Hoai},
  journal={Computers, Materials, \& Continua},
  volume={82},
  number={2},
  pages={2703},
  year={2025},
  publisher={Tech Science Press}
}

@article{zhang2026gc,
  title={GC-ViT: Graph Convolution-Augmented Vision Transformer for Pilot G-LOC Detection Through AU Correlation Learning},
  author={Zhang, Bohuai and Xu, Zhenchi and Li, Xuan},
  journal={Aerospace},
  volume={13},
  number={1},
  pages={93},
  year={2026},
  publisher={MDPI}
}

@article{venkatraman2025sag,
  title={SAG-ViT: a scale-aware, high-fidelity patching approach with graph attention for vision transformers},
  author={Venkatraman, Shravan and Walia, Jaskaran Singh and Joe Dhanith, PR},
  journal={Complex \& Intelligent Systems},
  volume={11},
  number={10},
  pages={428},
  year={2025},
  publisher={Springer}
}

@article{dogga2026hybrid,
  title={Hybrid vision transformer and graph neural network model with region-adaptive attention for enhanced skin cancer prediction},
  author={Dogga, Aswani and R, Sivasubramanian and S, Shanthi},
  journal={Scientific Reports},
  volume={16},
  number={1},
  pages={5385},
  year={2026},
  publisher={Nature Publishing Group UK London}
}

@article{jia2025multimodal,
  title={Multimodal depression detection based on an attention graph convolution and transformer},
  author={Jia, Xiaowen and Chen, Jingxia and Liu, Kexin and Wang, Qian and He, Jialing and Jia, X and Chen, J and Liu, K and Wang, Q and He, J},
  journal={Mathematical Biosciences and Engineering},
  volume={22},
  number={3},
  pages={652--676},
  year={2025},
  publisher={American Institute of Mathematical Sciences}
}

@inproceedings{fixelle2025hypergraph,
  title={Hypergraph vision transformers: Images are more than nodes, more than edges},
  author={Fixelle, Joshua},
  booktitle={Proceedings of the IEEE/CVF Conference on Computer Vision and Pattern Recognition},
  pages={9751--9761},
  year={2025}
}

@article{kim2024rethinking,
  title={Rethinking attention mechanisms in vision transformers with graph structures},
  author={Kim, Hyeongjin and Ko, Byoung Chul},
  journal={Sensors},
  volume={24},
  number={4},
  pages={1111},
  year={2024},
  publisher={MDPI}
}

@article{chen2024unified,
  title={A unified and biologically plausible relational graph representation of vision transformers},
  author={Chen, Yuzhong and Xiao, Zhenxiang and Du, Yu and Zhao, Lin and Zhang, Lu and Wu, Zihao and Zhu, Dajiang and Zhang, Tuo and Yao, Dezhong and Hu, Xintao and others},
  journal={IEEE Transactions on Neural Networks and Learning Systems},
  volume={36},
  number={2},
  pages={3231--3243},
  year={2024},
  publisher={IEEE}
}

@inproceedings{rashid2026shapbpt,
  title={ShapBPT: Image Feature Attributions Using Data-Aware Binary Partition Trees},
  author={Rashid, Muhammad and Amparore, Elvio G and Ferrari, Enrico and Verda, Damiano},
  booktitle={Proceedings of the AAAI Conference on Artificial Intelligence},
  volume={40},
  number={30},
  pages={25099--25107},
  year={2026}
}

@article{devaguptapu2024semantic,
  title={Semantic Graph Consistency: Going Beyond Patches for Regularizing Self-Supervised Vision Transformers},
  author={Devaguptapu, Chaitanya and Aithal, Sumukh and Ramasubramanian, Shrinivas and Yamada, Moyuru and Kaul, Manohar},
  journal={arXiv preprint arXiv:2406.12944},
  year={2024}
}

@article{cascone2023classification,
  title={Classification of fragments: recognition of artistic style},
  author={Cascone, Lucia and Nappi, Michele and Narducci, Fabio and Russo, Sara Linda},
  journal={Journal of Ambient Intelligence and Humanized Computing},
  volume={14},
  number={4},
  pages={4087--4097},
  year={2023},
  publisher={Springer}
}

@inproceedings{suvorov2022resolution,
  title={Resolution-robust large mask inpainting with fourier convolutions},
  author={Suvorov, Roman and Logacheva, Elizaveta and Mashikhin, Anton and Remizova, Anastasia and Ashukha, Arsenii and Silvestrov, Aleksei and Kong, Naejin and Goka, Harshith and Park, Kiwoong and Lempitsky, Victor},
  booktitle={Proceedings of the IEEE/CVF winter conference on applications of computer vision},
  pages={2149--2159},
  year={2022}
}
 
\end{document}